# 3D Convolution Neural Network based Person Identification using Gait cycles

Ravi Shekhar Tiwari[1], Supraja P[2,*], Rijo Jackson Tom[3],

[1,2]School of Computing, SRM Institute of Science and Technology, Kattankulathur, India

[3]Associate Professor, Department of computer science, CMR Institute of Technology Bangalore, India

Email: tiwari11.rst@gmail.com[1], p.supraja18@gmail.com[2,*] rijojackson@gmail.com[3]

*Abstract*—**Human identification plays a prominent role in terms of security. In modern times security is becoming the key term for an individual or a country, especially for countries which are facing internal or external threats. Gait analysis is interpreted as the systematic study of the locomotive in humans. It can be used to extract the exact walking features of individuals. Walking features depends on biological as well as the physical feature of the object; hence, it is unique to every individual. In this work, gait features are used to identify an individual. The steps involve object detection, background subtraction, silhouettes extraction, skeletonization, and training 3D Convolution Neural Network (3D-CNN) on these gait features. The model is trained and evaluated on the dataset acquired by CASIA -B Gait, which consists of 15000 videos of 124 subjects' walking pattern captured from 11 different angles carrying objects such as bag and coat. The proposed method focuses more on the lower body part to extract features such as the angle between knee and thighs, hip angle, angle of contact, and many other features. The experimental results are compared with amongst accuracies of silhouettes as datasets for training and skeletonized image as training data. The results show that extracting the information from skeletonized data yields improved accuracy.**

## I. INTRODUCTION

ARTIFICIAL INTELLIGENCE (AI) is used in fields like healthcare, network security, surveillance, identification, biometrics, and many others [1]. Even our phones, laptops, desktops, and numerous things around us have become a part of these technologies which enable us to utilize them in an efficient way. Machine Learning is a part of artificial intelligence, which makes the machines to learn by giving examples and perform actions based on similar values [2]. There are several supervised and unsupervised algorithms which can be used for prediction and automation. Deep Learning or deep structured learning allows machine learning to move closer to its original goals [3]. The main motive of AI is to automate things and make it simpler to humans. Almost 50% of the applications we use are based on AI viz. Microsoft Cortana, Apple Siri, etc., they are artificial intelligence agents who help us to attain our objectives.

Biometric is related to the human characteristic, and it plays a massive role in human identification. Biometric is used for access control and human identification in most of the scenarios. This type of recognition techniques helps us to identify the individuals by their various physical as well as identical biological features. Face recognition is a technique which allows us to identify the humans on the basis of their facial features at a certain point of time [4] this identification framework can be effectively infiltrated by manipulating facial features as well by identical persons.

Researchers have proved that fingerprint is one of the most secure ways to identify individuals till now and has been used extensively in many places [5], but due to the technology innovation, there are several incidents where fingerprint identification have been fooled [6] by using 3D printed silicon mounds of the victims. New approaches, which study the iris pattern known as iris recognition system, is potable and is being used in industries as well. The main drawback in this is that it cannot differentiate between living and dead also, it can be triggered by high-quality photographs [7]. The advantages of the biometric system are that they cannot be fooled easily and the person needs to be a perfectionist in order to fool these systems if any little dissimilarity is present then culprit has to face the consequences. Generative Adversarial Networks (GANS) is being used to generate videos as well as the sound of the targeted individuals.

Gait is defined as the study of human locomotion. Researchers have proved that every individual has identifiable walking style, i.e. gait cycle which solely depends on their physical as well as their biological factor such as bone-density, bone-weight, height, weight, and many other features respectively [8]. Gait analysis is a medical term, earlier it was used to detect gender [9], comparison of gait analysis between man (different age) how their walking pattern differs with the age [10] kinematics in athletes at which angle the athlete will gain more momentum, line, and angle of action [11] as a cue to identify the person by way they walk [12]. Since every individual has a unique and identifiable gait cycle -it is a game-changer as it does not intimidate the targeted person that they are watches also gender classification can be done [8]. Researchers have proved that gait cycle is also unique features similar to an iris scan, fingerprint scan and facial recognition [13] and can be used to identify human from a distance and have several disadvantages over traditional existing biometric recognition system [8]. Various parts of the human body show unique features which can be used to identify the individuals [14]. Gait cycle is divided into two phases stance phases which is further divided into four sub-phases- Initial Contact, Loading Response, Mid-stance, Terminal-stance and swing phase, which is also subdivided into four sub-phases, i.e., Pre-swing, Initial swing, mid swing, terminal swing in ratio 60:40, [15]. With the help of these phases, humans are able to walk in their unique way [16] with the coordination of muscle in loading and unloading off weight from one leg to another and several biological sequences of actions.

The video or sequence is passed through the object detector, which detects the presence of human and tracks the human [17] the extracted frames background is subtracted, and the silhouette is extracted. Silhouettes are a collection of features from the gait cycle of the individual; it has numerous information with respect to the individual [18]. The thin skeleton-like structure [19] were obtained from these silhouette by applying pixel reducing algorithm on silhouettes gave the different gait cycle quantities features primarily from lower body part [20]. These videos were propagated to the 3D-CNN to extract the stance features from the pixel-reduced silhouette, and it was forward propagated to the fully connected layer. We have the ability to predict by analyzing the walking style of individual [8], so artificial intelligence can also be trained to indent the individual by analysis their gait cycle [3]. Deep learning plays a very crucial role here as it has the ability to learn by going through previous learning data and used its memory to identify or perform a specific action basis of the previous training dataset. Gait is one of the ways of identification of individuals so researchers have more focus on this biometric identification technique as it does not need target cooperation and it can recognize individuals from a safe distance [12] also it can be used to surveillance the targeted individual without intimating them [14].

The contributions of the paper are as follows:

a. The video sequences are analyzed for person detection, and the silhouette is extracted, and thinning skeletonization is performed, and the dataset is generated for training the 3D-CNN.

b. 3D-CNN , for both silhouette based dataset and skeletonization based dataset and the accuracy, is compared amongst them.

The rest of this paper is organized as follows. Section II discusses the related works. The 3D CNN implementation for gait analysis is detailed in section III. Section IV shows the results obtained and the discussions that are followed. Section V concludes the work.

## II. RELATED WORK

Gait recognition is one of the latest technologies, which is being used as a biometric trait to identify the human using silhouettes or forms of silhouettes to identify human [19]. Silhouettes are a source of enormous information which can be used to identify the human by their gait cycles [18]. Gait recognition is one of the most secure ways to identify individuals [20].

Object Detection and tracking are the two most vital part of video analysis in surveillance. It includes the non-overlapping techniques in multi-camera scenarios to detect humans and track them. Another system proposed uses the general image information extracted from the field view to increase the accuracy to detect the object and track it [21]-[24]. It is expensive as camera, and environment setup is expensive. This is the initial step of any surveillance system to detect the presence of an object in the frame. Background modulation is the next step taken in a surveillance system to minimize the noise in the frames and to extract predefined object's features with less interference with surrounding and for better pre-processing. Mean shift (MS) segmentation algorithm is used to create the boundary around the object by mask obtained from the adaptive Gaussian mixture modeling by minimizing the noise present in the frame. Tracking object can be segmented into two phases 1) Intra -camera tracking (tracks object within one camera) and 2) inter-camera tracking (tracks objects using more than one camera) [25]. Mean Shift is one of the most popular techniques because of simplicity in applying the concept and effectiveness it just keeps solitary speculation and uses the angle of the information conveyance in looking for the most extreme conceivable applicant. Thus, it is exceptionally computationally proficient. Notwithstanding, regular MS tracker is inclined to losing tracks because of the quick development of the object, the effectiveness of the mean shift algorithm decreases when multiple objects are present. So here we are implying the mean shift algorithm which cascade to detect the multiple objects from the frames and create a bounding box while tracking the object. We are applying a filter to minimize the interference from the surrounding and to clean each frame before detecting the object, so the accuracy is high in all the scenarios. All the cascades features allow us to detect various objects simultaneously in frames.

Background Subtraction (BG) is the first preprocessing step in video surveillance. It subtracts the background and extracts the foreground i.e. it aims to separate foreground from stationary background objects [26]. BG is used to differentiate between stationary and non-stationary objects in video frames. It has several steps like first frame initialization, background modeling, minimizing the noise, and set off frames [27]. Background modeling is the most important process which enables us to convert the extracted frame in a predefined format for pre-processing, and it is used to detect the moving objects in frames. The implied algorithm is to subtract the frame in which the object is not present from the next consecutive frames, applying the filters to further minimize the noise by applying erosion and dilation [28] and applying the threshold according to the importance of the real-time computations of the surveillance system. The BS algorithm is improved and has become much easier to implement and time taken to reduce the frame has been reduced, thus enabling us to implement in real-time scenarios. The main challenge in BS is like illumination, shadows of non-stationary objects, as well as stationary, external noise and many other things earlier it takes huge computation as well as the long time to process each frame but researchers, have found methods to make it cost-efficient and easy to implement [29]. There are many BS algorithms, but the best one takes less time to implement, and it can be implemented is any condition irrespective of light intensity [30]. A robust system was developed for virile tracking and classification was proposed and tested under a harsh condition such as a rainy day and sunny day [31]. As of recent wide scope of BS calculations have been created to enhance their execution. The movement-based perceptual gathering a spatial-transient saliency calculation is proposed to perform BS and it is relevant to scenes with exceptionally unique foundations.

Pixel reduction algorithms play a very crucial role in pattern recognition. It is a preprocessing technique to identify different types of images of different objects. Pixel reduction algorithm is objective is to reduce the duplicated pixels value around the binary images or video frames, thus evolving it into a skeleton-like structure, which will be used to identify the objects. Pixel reduction algorithm plays an important role in many applications such as fingerprint recognition, classification, medial application, and many other fields due to its huge importance [32]. The criteria for the pixel reducing algorithm to work are to choose $m*n$ neighborhood chooses around the currently selected pixel, if the currently selected pixel satisfies the rule, then the surrounding pixel is deleted. This deletion is based on the iteration over the surrounding pixels [33]. Good pixel reduction algorithm should preserve the connectivity, and secondly, it should produce a single pixel thick width last, but most important is that it should preserve the shape of the object [34]. From this pixel reduction algorithm, we can extract features from the image and can be used to classify or identify the object with high accuracy. It also emphasis on removing the unwanted object by applying threshold and reduce the noise which originates from the shadow or due to the motion of the objects.

Gait cycle is referred to as the study of semantic of human locomotion. Gait cycle consists of two-phases 1) stance phase and 2) swing phase in ratio 60:40. These phases are further divided into eight different phases 1) Initial Contact 2) Loading Response 3) Mid-Stance 4) Terminal Stance 5) Pre-Swing 6) Initial Swing 7) Mid-Stance 8) Terminal Swing and with the help of coordination of different muscles human are able to walk. Gait is an identical feature of human as same as the fingerprint scanner, iris scanner, and face recognition. Researchers have shown that gait can be used to identify the human by the help of their gait cycle. Researches have categorized gait recognition as model-based [35] and appearance-base [36]. Researchers have collected various features from the silhouette of the human, which has unique modeling signatures through the video frames features stored in the form of angles, height. Researchers have also proposed a method which identifies the human with the help of gait energy images (GEI) and training the CNN to identify the human by considering cross-section and cross walking pattern with an average recognition rate of 94% [37]. Studies show that GEI and silhouette have their own stability, simplicity, and uniqueness so it is been applied to human identification [38]. In this paper we have extracted the silhouettes from the video of an object after object detection and background subtraction, we further preprocessed these silhouettes to obtain thin fine line by using pixel reducing algorithm to focus more on lower body part features as well as to focus upper body part also.

Deep Learning Models are can learn with the help of training data with increment in the accuracy by building high-level features [39]. The convolution neural systems (CNNs) are a kind of profound models in which trainable channels and nearby neighborhood pooling activities are connected alternating on the crude information pictures, bringing about a chain of importance of progressively complex highlights [40]. It has appeared, when prepared with proper regularization CNNs can accomplish predominant execution on visual article acknowledgment assignments. Moreover, CNNs have been appeared to be invariant to specific varieties, for example, posture and lighting. CNN's have been principally connected on 2D pictures, to viably fuse the movement data in video examination, we propose to perform 3D convolution in the convolution layers of CNNs with the goal that discriminative highlights along both the spatial and the fleeting measurements are caught [41]. 3D CNN network architecture that creates various channels of data from nearby video outlines and performs convolution and subsampling independently in each channel. The last component portrayal is acquired by joining data from all channels.

## III. 3D-CNN BASED GAIT ANALYSIS

This work focuses on developing a biometric system for the purpose of security in restricted zones using Gait patterns of individuals. The strength of other biometrics like fingerprint and iris are known but still have been spoofed. Gait feature of humans can bring in an added security feature. The system works on identifying the moving person in the video and study the gait using 3D-CNN to identify the individuals. It is a combination of several individual algorithms associated together in such a way that it gives one of the best procedure to identify the individual by their gait cycle. The algorithms are object detection (human detection and tracking), background subtraction, silhouette extraction, silhouette pixel reduction, and passing the reduced image through trained 3D-CNN, Fully-connected layers and finally to the classifier to identify the individual. The stream of the procedure is shown in Fig 1.

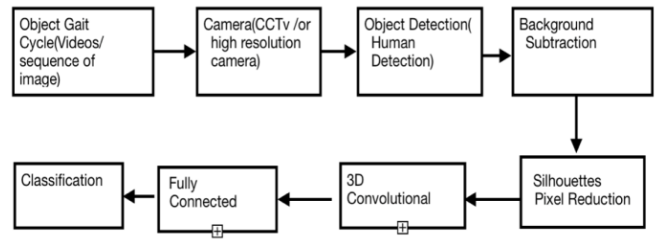

Fig 1. Steps involved in 3D-CNN based gait analysis for person identification.

### A. Object Gait Cycle (Videos /Sequence of Images)

The work uses the Casia-B Gait dataset [44] [45], which consist of 124 individuals captured from different 11 angles, were captured and named accordingly. The objects are walking with and without the external object and the three parameters, edge, attire and conveying condition changes, were independently considered while capturing the dataset. Gait cycle consists of two phases-stance phases and swing phase in ratio 60:40 [42] these stance phase and swing phase is divided into 8 sub-phases-Initial Contact, Loading Response, Mid-stance ,Terminal-stance, Pre- swing, Initial swing, mid swing, terminal swing [43]. Fig. 2 represent the stance phase and swing phase of a human gait cycle in ratio 60:40. It shows a person walking pattern as the coordination of these identical phases, which is being used to identify the individual. These patterns are almost impossible to capture as it depends on the biological and physical factor of the individuals.

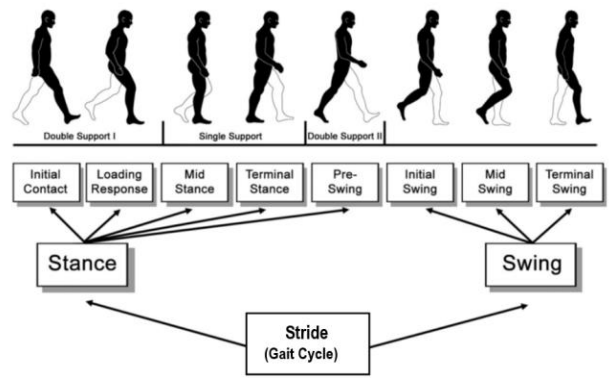

Fig. 2 Gait Cycle

### B. Human detection and tracking

For the evaluation purpose, we are using normal camera which has a range up to 40 meters (approx.) are used to capture the gait cycle of the targeted person and hence to identify them in real-time scenarios. The ongoing presentation of profundity cameras has empowered the catch of 3-D joint areas without the utilization of markers. This data has been utilized in real life and stride acknowledgment applications. Gait cycle is dependent on physical as well as genetically feature; object cooperation is not required this identification system. From the videos captured by the camera we are detecting the presence of human and tracking it. The human detection and tracking work on localizing the frames of the video then detecting the human and tracking it in the given video frame as to avoid any other living organism detection and giving false results. It increases the performance and throughput time because frames in which object is not detected is discarded.

### C. Background Subtraction

Background subtraction is one of the normal methods to recognize and find the moving article in a sequence of pictures [37]. It reduces the size of the image as well as the convert sequence of images into black and white channel from red, blue, green i.e., 3 channel [40],[41] as a result we get sequence of frame which is having our object (human) as white pixel and rest is all black i.e., silhouettes obtained from the sequence of images. Silhouettes are one of the best ways to extract the qualitative as well as quantitative unique information about the individuals, and hence, it can be used to study the gait cycle of the individual.

### D. Silhouettes Pixel Reduction

Pixel reduction is an algorithm that reduced the density of the pixel by taking consideration of mathematical formulae by grouping pixels into super-pixels defined as a group of a pixel arranged into the logical meaningful atomic region which can be used to reduce the redundancy of the pixel. Pixel reduction helps us to create skeleton-like structure from the obtained silhouettes also the size of the dataset is reduced. These skeleton-like structures are being used to extract the various angle between different bones in object lower body part.

### E. 3D-Convolutional Layer

Till now, only 2D-CNN has been applied to analyze the gait cycle of the individuals. In this paper, we will apply 3D-CNN to analyze the gait cycle of the individual and identify the individual in real-time. We will feed the pixel reduced video to the 3D Convolution. It accepts input as a sequence of frames which is adjacent to each other to extract the feature of each individual.

The 3D convolution is accomplished by convolving a 3D part to the 3D shape shaped by stacking different adjacent casings together.

$$v_{ij}^{xyz} = tanh\left(b_{ij} + \sum_{m}\sum_{p=0}^{P_i-1}\sum_{q=0}^{Q_i-1}\sum_{r=0}^{R_i-1} w_{ijm}^{pqr} v_{(i-1)m}^{(x+p)(y+q)(z+r)}\right) \qquad (1)$$

Equation (1) represents the 3D-Convolution where the $R_j$ is the size of the 3D kernel along the temporal dimension $w_{ijm}^{pqr}$ is the $(p,q,r)^{th}$ value of the kernel connected to the m$^{th}$ feature map in the previous layer.

### F. Fully Connected Layer and Classification

The fully connected layer is like how neurons are arranged in a conventional neural system. Along these lines, every hub in a completely associated layer is straightforwardly associated with each hub in both the past and in the following layer. It has the final layer which has activation function sigmoid that acts as a classifier. The sigmoid formulae are as follows:

$$S_i = \eta\left(\varphi\big(\varphi(x), \varphi(x_i)\big)\right) \qquad (2)$$

where, $x$ is a diving GEI, $x_i$ is an arcade GEI, $\varphi$ ventures $x$ and $x_i$ into an acknowledged space, $\varphi$ processes the flourishing abnormality in the midst of its two information sources, and $\eta$ predicts the last likeness. The softmax function shows the probability of each neuron what has been activated based on the previous layer.

### G. Dataset

The dataset on which the 3D CNN had been trained is acquired from the Institute of Automation,Chinese Academy of Sciences (CASIA). Dataset-B is a large dataset consist of 124 objects. The dataset information is pictured below in Table 1:

TABLE I OVERVIEW OF DATASET

(* MM: Walking Status, NM:Normal Walking, CL:Walking with Coat, BG: Walking with Bag. The dataset consist of approximately 15000 videos of each object in 11 different angles.)

| No of Persons | View Angle(in degrees) | Walking Status (MM) | No of videos |
|---|---|---|---|
| 01-30 | 0,18,36,54,72,90,108,126,144,162,180. | MM,NM,CL,BG | 121(each) |
| 31-60 | 0,18,36,54,72,90,108,126,144,162,180 | MM,NM,CL,BG | 121(each) |
| 61-90 | 0,18,36,54,72,90,108,126,144,162,180 | MM,NM,CL,BG | 121(each) |
| 91-124 | 0,18,36,54,72,90,108,126,144,162,180 | MM,NM,CL,BG | 121(each) |

### H. System Specification and Framework

We trained this neural network on the Dell workstation with specifications of RAM: 128 GB DDR4, NVIDIA Quadro M2000 4 GB, 14 cores, 35 MB catch and 12 TB of storage. The frameworks used in this work are Keras, Tensor Flow, Theano, OpenCV, Scikit-learn, and various supporting libraries. The 3D CNN trained on this system for 7 consecutive days by preprocessing 124 objects walking dataset and attain the following parameters while training and testing on the dataset by applying the object detection, Background Subtraction and Silhouettes Extraction, Pixel Reduction Algorithm. The obtained data is formulated below.

*I.  Preprocessing Of Dataset and Network Architecture*

In this paper, we preprocessed the dataset by using Scikit-learn library and OpenCV and various supporting libraries. First, we detected the object by applying the object detection algorithm, if the object is present in the frame then we processed it further by subtracting the background, silhouettes extraction and applying the pixel reduction algorithm to obtain a thin skeleton-like structure as mentioned in Fig 1.The full flow of the process of training and preprocessing the dataset is shown in Fig.3

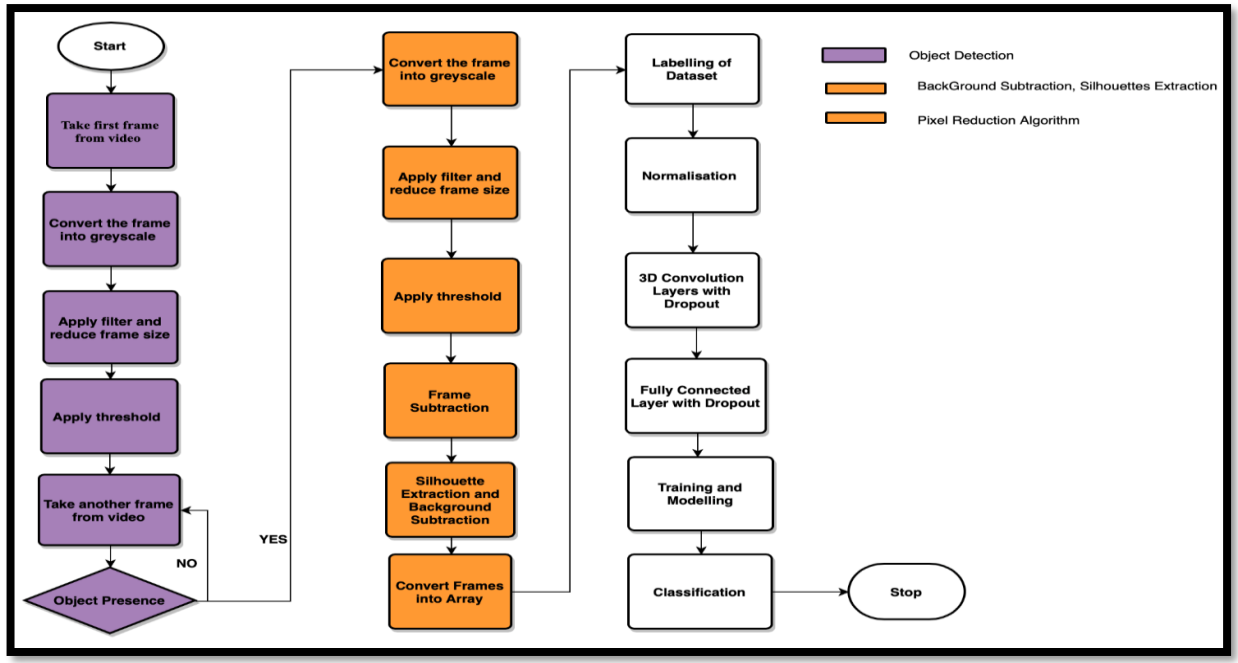

Fig 3. Flowchart of the steps involved in proposed model

The full process of the experiment is described training and the network architecture. As shown in Fig. 3 the first preprocessing step is to determine the presence of an object in the frame if the object is present then it will be converted into the greyscale, noise is reduced by applying the filter, and the size of the frame was reduced. Then the remaining noise was removed by applying the filter and then the background was subtracted from the frame silhouettes were obtained. On these obtained silhouettes the pixel reducing algorithm was applied to remove the redundant pixel and obtain the skeleton structure of the object. The skeleton of the object was feed into the 3D CNN with dropout layer after normalization which is connected to the fully connected layer with dropout to reduce the overfitting. The below fig. 4 shows the preprocessed frames of the dataset.

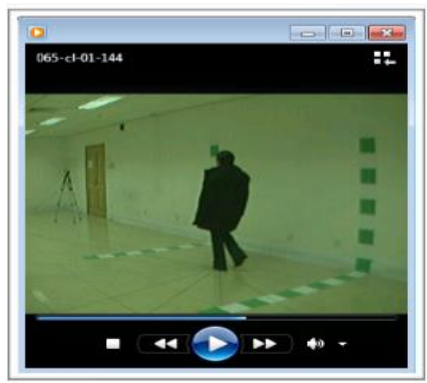

Fig. 4 Frame of the dataset

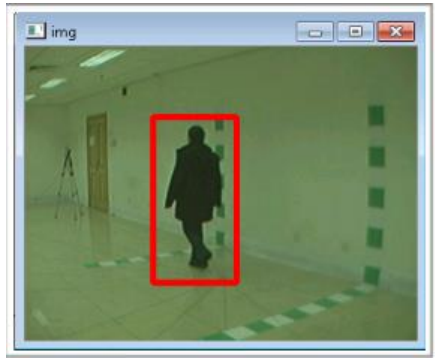

Fig. 5 Object detected in the frame

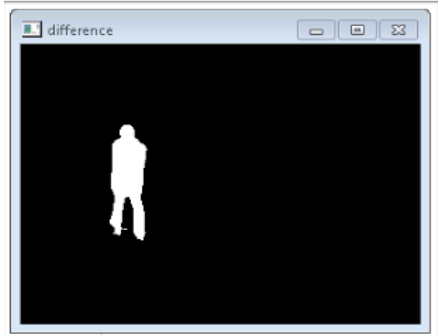

Fig. 6 Background Subtraction and Silhouettes Extraction

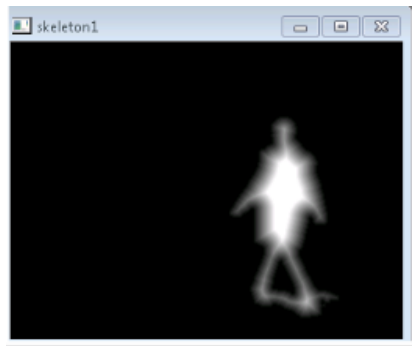

Fig. 7 Partial Skeleton

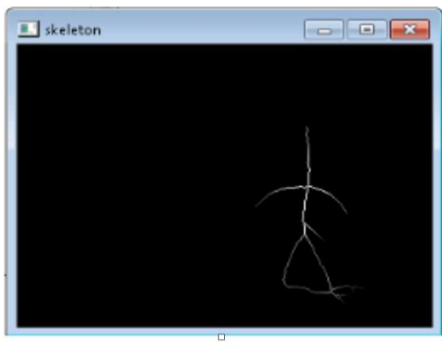

Fig. 8 Skeletonise form of Object

The above figure represents the various stages of preprocessing, in each step the frame of video is subjected to various algorithms as shown in Fig. 3. The Fig.5 represent the object detection in the video as to determine the presence of the object in frame as to process it further. The Fig. 6 represents the background subtraction and silhouettes extraction, finally the Fig.7 and Fig. 8 represents the pixel reduction algorithm applied on the extracted silhouette in Fig. 6 as to obtain various qualitative and quantitative feature of the object. In this system, we have proposed a human identification with skeletonization of its gait cycle 3D CNN.

*J. Training and Testing*

This network was trained on the dataset obtained from the CASIA Gait-B which consist of 124 objects. The dataset was divided into a two-part training set and test set in a ratio of 80:20. The flow of the process in the training set is explained in Fig. 9. We obtained an accuracy of 94.27% using thinned images for training and an accuracy of 90.56% by using silhouette. Fig. 10 shows the output of skeletonization algorithm trained in the 3D CNN.

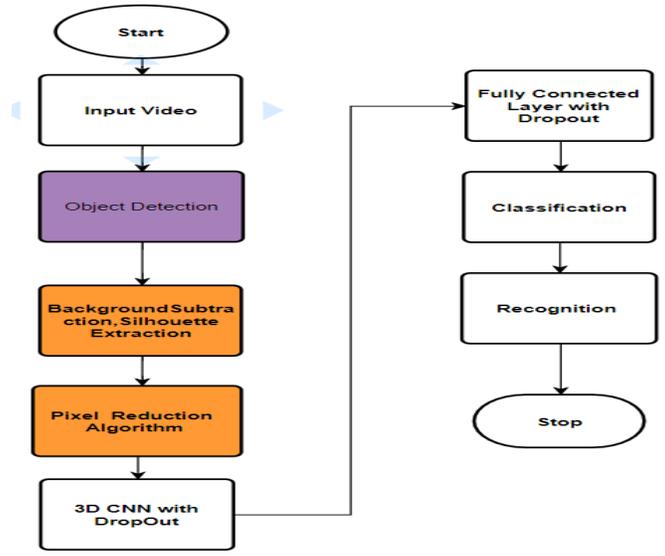

Fig. 9 Flow of process for testing the trained network

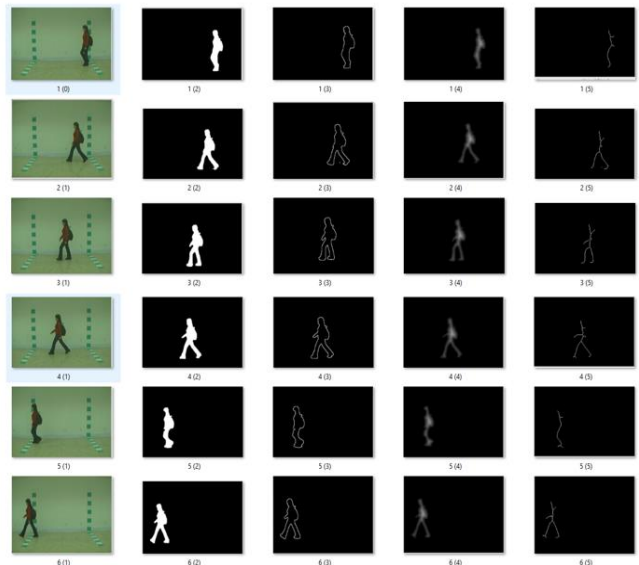

Fig. 10 The output of skeletonization algorithm for generating dataset for training the neural network.

## IV. RESULTS AND DISCUSSION

We training of the network took seven days on the dataset obtained from the CASIA Gait-B with the hardware specified in section III,H. The following results were obtained as shown in Table II. The graphs obtain the training set and on the test set over 100 epochs are pictured below.

On training the 3D CNN on the dataset provided by Casia, we obtain the accuracy and error graphs which are shown in Fig. 11 and Fig.12. The obtained accuracy is 94.27% on dataset using the thinning algorithm and accuracy is 90.56% on using silhouette.

The Fig. 13 represents the mean absolute error and categorical accuracy respectively which is 0.073% and 94.27%.. The Fig. 14 represents the loss model obtained on training set is 1.856%.

TABLE II COMPARISON TABLE PREPROCESSED WITH SILHOUETTES AND PIXEL REDUCTION ALGORITHM

| Parameters | Silhouette based system accuracy (in %) | Skeletonized image based system accuracy (in %) |
|---|---|---|
| Accuracy | 90.16 | 94.27 |
| Loss | 0.2980 | 1.856 |
| Value Accuracy | 79.84 | 90.56 |
| Categorical Accuracy | 90.16 | 94.27 |
| Mean Absolute Error | 0.0131 | 0.073 |

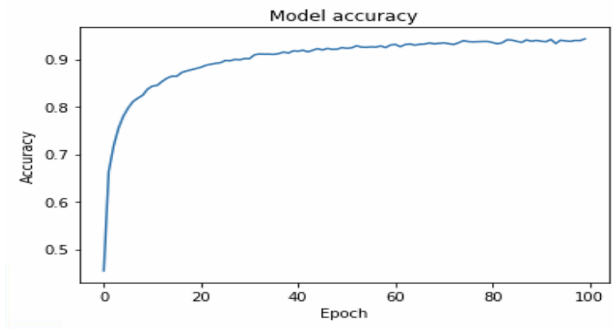

Fig 11 Trained model Accuracy

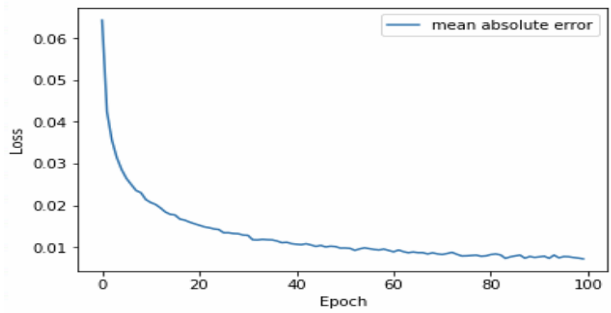

Fig 12. Trained model error

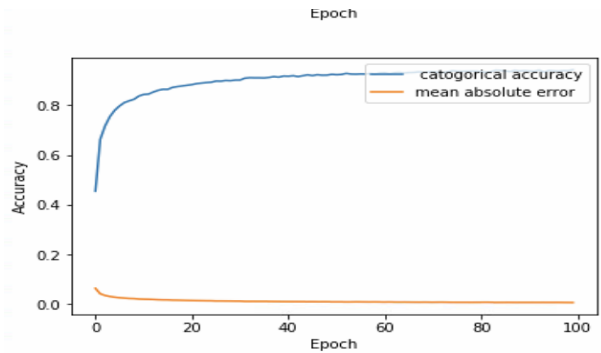

Fig.13 Categorical Accuracy and Mean Absolute ErroR

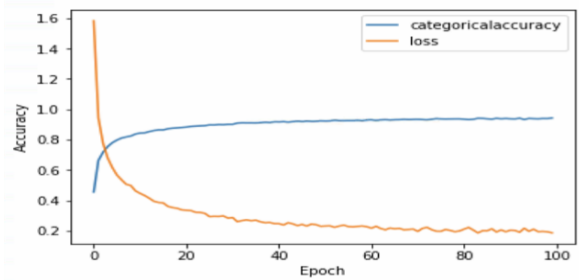

Fig.14 Categorical Accuracy and Loss

## V. Conclusion

We developed 3D CNN model to identify the individuals from their Gait Cycle. This model used structural as well as the physical dimension by carrying out 3D-CNN. The constructed deep CNN generate numerate numerous features by adjacent frames and perform 3D-Convolution and max pooling. The frames are preprocessed in form of skeleton like structure by preprocessing the original dataset in order to obtain features to measure the various angles mainly in lower body part such as femur and tibia, calcareous and surface of contact, hip angle and many more. Since theses angles are solely dependent on biological as well the physical features and these features are unique. The final features obtained are combination of the all features from all channels. We trained the model on Casia Gait dataset-B consists of 124 objects walking pattern from 11 angles. We preprocessed each video of object as shown in Fig. 1 then propagated it to the fully connected layer the types of pixel reduction algorithm used are normal skeleton, gradient, medial skeleton, thinning, and silhouettes. The values obtained from theses pixel reducing algorithm after training and testing were tabulated and compared as shown in Table 2. We attained comparable accuracy hence it can be applied to the real world scenario.

### Acknowledgments

In this paper, we used supervised 3D-CNN to identify the person by analyzing their gait pattern. There are existing networks which achieved satisfactory performance in analyzing the object by their gait cycle. This 3D-CNN model
was trained using supervised algorithm which requires labelled samples to train and test, whereas it can be reduced using unsupervised algorithm. There are an obvious number of improvements, which can be made in the future, includes Improvement of Pattern Classifiers, Upgrading display fitting calculations. Also this project restricted to procuring stride marks from a solitary camera position, it is beneficial to be able to perceive individuals from various edges, instead of being constrained to only one edge.